\documentclass[letterpaper, 10 pt, conference]{ieeeconf}  % Comment this line out if you need a4paper

\IEEEoverridecommandlockouts                              % This command is only needed if 
\usepackage{graphicx} % for pdf, bitmapped graphics files
\usepackage{amsmath} % assumes amsmath package installed
\usepackage{amssymb}  % assumes amsmath package installed
\usepackage[group-separator={,}]{siunitx}
\usepackage{xcolor}

\title{\LARGE \bf
Efficient and Diverse Generative Robot Designs using Evolution and Intrinsic Motivation
}

\author{Leni K. Le Goff$^{1}$ and Sim\'on C. Smith$^{1}$% <-this % stops a space
\thanks{$^{1}$School of Computing, Engineering \& The Built Environment, Edinburgh Napier University, UK.
        {\tt\small l.legoff2@napier.ac.uk}, {\tt\small s.smith2@napier.ac.uk}}%
}

\begin{document}

\maketitle
\begin{figure*}[t!]
  \centering
  \includegraphics[width=\linewidth]{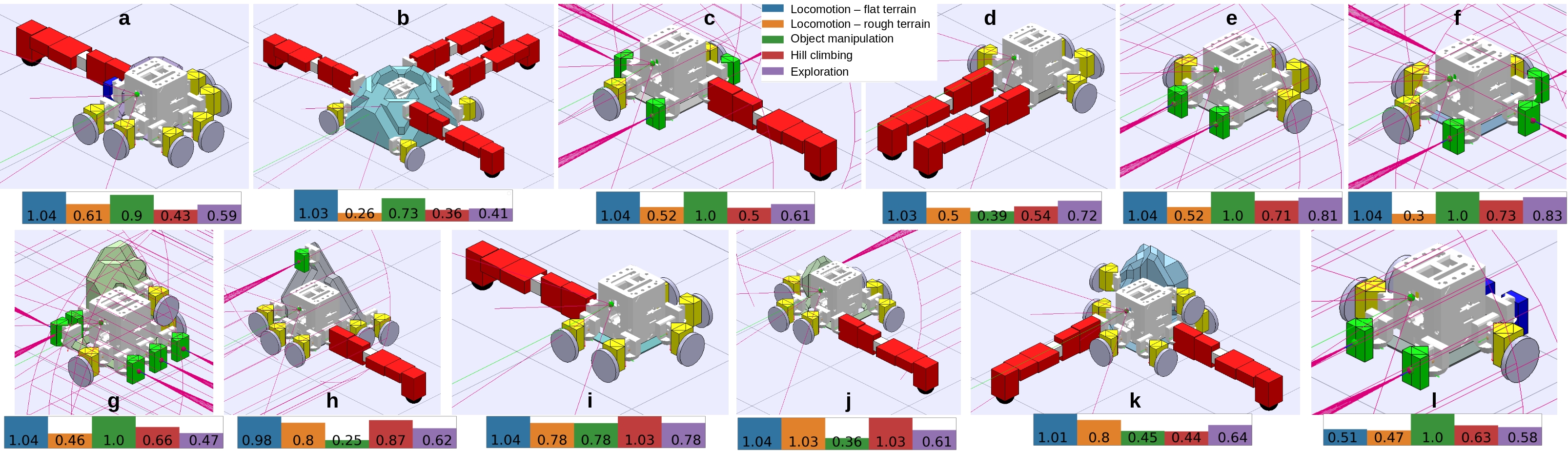} 
  \caption{Robots generated by our approach MEHK. Exploration values refer to the evaluation phase during morpho-evolution. The other values correspond to the downstream task scores.}
  \label{fig:rob_ex}
\end{figure*}
\thispagestyle{empty}
\pagestyle{empty}

%%%%%%%%%%%%%%%%%%%%%%%%%%%%%%%%%%%%%%%%%%%%%%%%%%%%%%%%%%%%%%%%%%%%%%%%%%%%%%%%
\begin{abstract}

Methods for generative design of robot physical configurations can automatically find optimal and innovative solutions for challenging tasks in complex environments. The vast search-space includes the physical design-space and the controller parameter-space, making it a challenging problem in machine learning and optimisation in general. Evolutionary algorithms (EAs) have shown promising results in generating robot designs via gradient-free optimisation. Morpho-evolution with learning (MEL) uses EAs to concurrently generate robot designs and learn the optimal parameters of the controllers.
Two main issues prevent MEL from scaling to higher complexity tasks: computational cost and premature convergence to sub-optimal designs. To address these issues, we propose combining morpho-evolution with intrinsic motivations. Intrinsically motivated behaviour arises from embodiment and simple learning rules without external guidance. We use a homeokinetic controller that generates exploratory behaviour in a few seconds with reduced knowledge of the robot's design. Homeokinesis replaces costly learning phases, reducing computational time and favouring diversity, preventing premature convergence.
We compare our approach with current MEL methods in several downstream tasks. The generated designs score higher in all the tasks, are more diverse, and are quickly generated compared to morpho-evolution with static parameters.
\end{abstract}

%%%%%%%%%%%%%%%%%%%%%%%%%%%%%%%%%%%%%%%%%%%%%%%%%%%%%%%%%%%%%%%%%%%%%%%%%%%%%%%%
\section{Introduction}

The physical configuration of a robot, including its body, limbs, sensors and actuators, plays a crucial role in its capacity to solve any task. In domains where no robotic solution exists, experts can develop a robot, including its physical design and training of its controller. As another solution, generative artificial intelligence automatically designs robots and their controllers, accelerating the production of innovative digital and physical artefacts~\cite{epstein2023art,buonamici2020generative}. %,saadi2023generative}. 
%Human design, usually biased by biological ones, are able to solve a plethora of tasks given the correct training. However, these designs are sub-optimal and usually restricted by external constraints like hardware availability, reuse of the same robot for different tasks or price.
%The goal of automatic generative design applied to robotics is to find optimal physical robot configurations.
Examples of generative design applied to robotics include the use of reinforcement learning (RL)~\cite{luck2020data,li2024reinforcement}, topology optimisation~\cite{matthews2023efficient} and evolutionary algorithms (EAs)~\cite{gupta2021embodied,le2022morpho,jelisavcic2019lamarckian,li2023evaluation}.
%,sims1994evolving,lipson2000automatic,cheney2014unshackling,cheney2018scalable,kriegman2020scalable,miras2018search,hornby2003generative}. 
While RL and topology optimisation are limited to gradient-based design-spaces, gradient-free methods such as EAs offer more flexibility. The generation of robot designs using artificial evolution, \textit{morpho-evolution}, results in innovative configurations of actuators, sensors and passive parts able to solve complex tasks in dynamic environments. Examples of morpho-evolution include soft robotics~\cite{cheney2018scalable, mertan2024investigating}
%cheney2016difficulty,mertan2024towards}
, fine-tuning of existing designs~\cite{nygaard2018real} and the generation of a diverse set of high-performing designs~\cite{gupta2021embodied}.   

Classic morpho-evolution algorithms consider the solution-space as the joint space of the robot designs and controller parameters. 
%These approaches\cite{sims1994evolving,lipson2000automatic,cheney2014unshackling,cheney2018scalable,kriegman2020scalable,miras2018search,hornby2003generative} typically uses local variation (mutation and crossover) on the designs and controllers parameters to explore this joint space and find the optimal solution to solve a specific task by testing the robots against the environment and task.
Usually, these algorithms use EAs for robots design optimisation~\cite{sims1994evolving,lipson2000automatic,cheney2014unshackling}.
%,cheney2018scalable,kriegman2020scalable,miras2018search,hornby2003generative}.
However, these works study linear tasks in a reduced solutions-space, and the results are not applicable to physical robots. The design-space only includes proprioception and a reduced set of actuators, leaving out external sensors or actuators like manipulators.
%like infra-red sensors, cameras 

To scale morpho-evolution to more complex tasks and larger design-spaces integrating exteroception, the methods require a learning phase~\cite{eiben2020if,gupta2021embodied,le2022morpho}.
%,jelisavcic2019lamarckian,li2023evaluation}.   
These approaches combine learning and morpho-evolution.
%in a hierarchical architecture with two nested optimisation processes.  
First, an Evolutionary Computation (EC) algorithm generates robot designs, and a second process optimises the parameters of the controller for each design. In this paper, we refer to this framework as \textit{morpho-evolution with learning} (MEL).
 %In these approaches, the evolutionary process does not have pressure towards any downstream task. Only after a successful training of the controller, the process can asses the quality of the robot's morphology.
%However major issues prevent it to tackle real world problems.
%Current approaches of morpho-evolution include the optimisation of the robot’s physical configuration and the learning of a behavioural controller in a hierarchical architecture: \textit{morpho-evolution with learning} (MEL)\cite{gupta2021embodied,le2022morpho,jelisavcic2019lamarckian,li2023evaluation}. 
 %In general, these two processes are expensive as they require a large number of samples to converge. 

Two main issues prevent MEL from tackling higher-complexity tasks. First, 
%The design-space is intractable as it integrates actuators, sensors and their position in 2- or 3-dimensional space~\cite{gupta2021embodied,le2022morpho,buchanan2020bootstrapping}.
applying a learning phase to each design results in a high computational cost~\cite{luo2022effects}. For instance, authors in~\cite{gupta2021embodied} can generate diverse designs by combining deep RL and morpho-evolution. However, the approach requires a large computational power ($1152$ CPUs). %to explore $4,000$ designs.
Luo et al.~\cite{luo2022effects} compares morpho-evolution and MEL approaches, showing that with the same computational power, classic morpho-evolution (without learning) was able to run $200$ generations while MEL was still at its first generation.

Second, the studies~\cite{cheney2016difficulty,cheney2018scalable,mertan2024investigating} show that MEL framework suffers from premature convergence, resulting in locally optimal solutions reducing the diversity of possible designs. 
%In~\cite{cheney2018scalable}, the authors propose an \textit{innovation protection} mechanism to preserve novel designs.
The authors in~\cite{mertan2024investigating} show that the learning processes accentuate the premature convergence by comparing learning against a controller with static parameters, i.e. fixed controller. With a fixed controller, the diversity of design is higher than using a learning algorithm. This study is limited to distributed controller with only proprioceptive sensors.
%Thus, it is not clear if a fixed controller would have the same effect with a centralised controller including exteroception in the closed-loop control.
% Thus, increasing the diversity of designs.
% Novel design which are significantly different from previous generations will need time to adapt their controller and reach the best performance (\red{Im not sure about this sentence}). 
% In~\cite{mertan2024investigating}, the authors introduce a fixed controller architecture. The parameters of a fixed controller are set randomly at initialisation and are not updated, leading to higher diversity and performance. Meanwhile, traditional learning processes amplify the premature convergence.
% These studies are limited to proprioceptive controllers. 
% Thus, it is not clear if the innovation protection or a fixed controller would generate optimal and diverse solutions when including exteroception in the closed-loop control.
%work on a centralised controller such as neural network featuring sensors. %In the present study, we test the fixed control in such context.

%\subsection*{Intrinsically motivated morpho-evolution}
To address these issues, we propose to integrate intrinsic motivation with morpho-evolution to efficiently generate optimal and diverse designs.
Intrinsically motivated (IM) robots can behave in exploratory or curious ways without external rewards~\cite{ay2012guided}. The IM behaviour emerges from the robot's embodiment and internal representations of the world and its dynamics. Algorithms for IMs usually operate with fast online adaptation rather than lengthy learning phases. A robot without prior knowledge of the world's or its own dynamics starts exploring the environment in a few seconds.
IMs in robotics include guided RL, theories for internal model self-organisation, and automatic curriculum learning~\cite{liu2022learn,smith2011homeokinetic,smith2018evaluation,forestier2022intrinsically}.
Compared to MEL's learning phase, our approach can quickly decide if a design is viable by checking its IM behaviour. In this way, thousands of robots can be generated and evaluated in significantly less time than current MEL approaches.
%Another advantage of using IMs is that they require closed-loop control. 
%Our approach can operate with sensory information without losing fitness calculation speed. Other fast MEL approaches use CPGs to compensate for learning time. However, these approaches do not escalate to higher dimensions \red{what approaches uses CPGS? is this sentence correct?}.

To the best of our knowledge, only~\cite{hejna2021task} proposes a similar approach, task-agnostic morphology evolution (TAME). The authors combine morpho-evolution with empowerment~\cite{klyubin2005empowerment}. They show that empowerment is a good proxy for viable robot designs that generalise to multiple tasks. However, empowerment requires an accurate model of the dynamics between the robot and the environment. This model can be acquired as prior knowledge or learned during training. This last approach is inefficient for dynamic environments and requires an extensive data set for each robot.
%Also, TAME results did not include exteroception.

We propose to combine morpho-evolution with \textit{homeokinesis} (MEHK). Homeokinesis generates seemingly intelligent exploratory behaviour by adapting the controller's parameters in linear time by maximising predictability and sensitivity~\cite{der2012playful}. 
%Homeokinesis does not need a state space, it operates directly on the sensory inputs and actions.
%In a few seconds, a robot activates all its degrees of freedom in a coherent way. For example, a wheeled robot with a homeokinetic controller explores the environment by activating its actuators and reacting to the environment. When faced with a wall, the robot adapts its internal parameters to increase sensitivity, moving it away from the wall. The behaviour resembles a robot escaping a dead-end without needing a learning phase, external interference or repositioning. 
Homeokinesis replaces the lengthy learning process in MEL, e.g. RL, CMA-ES or EAs, required to test each generated design, with a short adaptation period that is able to make a robot move in an exploratory way in a few seconds of simulation.
MEHK calculates a fitness function as coverage of the environment. In this case, homeokinesis is a proxy towards optimal designs, filtering designs unsuitable for movement or interaction with the environment.
To test MEHK, we train the generated solutions in several downstream tasks. We select the best robots from a Pareto front between fitness and design diversity. We compare MEHK to a baseline of robots generated with morpho-evolution with a fixed controller.%, i.e. its parameters are static.   
%All the robots are trained and compared to a baseline which combines morpho-evolution with a fixed-controller.

%However, this approach raise another issue: \textit{how to evaluate if a design generated with morpho-evolution is functional?} 
%when the design space is large and expressive, i.e. can produce a large range of different kind of robots. 
%In the MEL framework, the learning process is used for optimising the controller to fit as best the current design and consequently evaluate its functionality. Without the learning process, one could apply the naive approach of a controller with a fixed architecture and parameters\cite{mertan2024investigating}. However, it is not guaranteed that a fixed controller would fit all the possible designs.  

%In our approach, we use spatial coverage arising from the self-organised exploration as the fitness for the robot's configurations. This fitness is fast to evaluate, as it does not require training. Thus, our approach quickly evaluates any configuration, compared to sample inefficient method like reinforcement learning or genetic algorithms.

Our contributions are: (1) introducing homeokinesis to the generative robotic design framework, including exteroceptive sensors; (2) introducing MEHK as a fast method for generating a high number of robot designs with low computational resources; 
%(3) MEHK does not collapse on local-minima designs, resulting in more extensive diversity of solutions when compared to other methods; 
and (3), we show that the solutions found by MEHK have higher performance for downstream tasks when compared to a morpho-evolution with a fixed controller.

\section{Methodology}

\subsection{Morpho-Evolution with Homeokinesis (MEHK)}\label{sec:mehk}

\begin{figure}[thpb]
  \centering
  \includegraphics[width=0.8\linewidth]{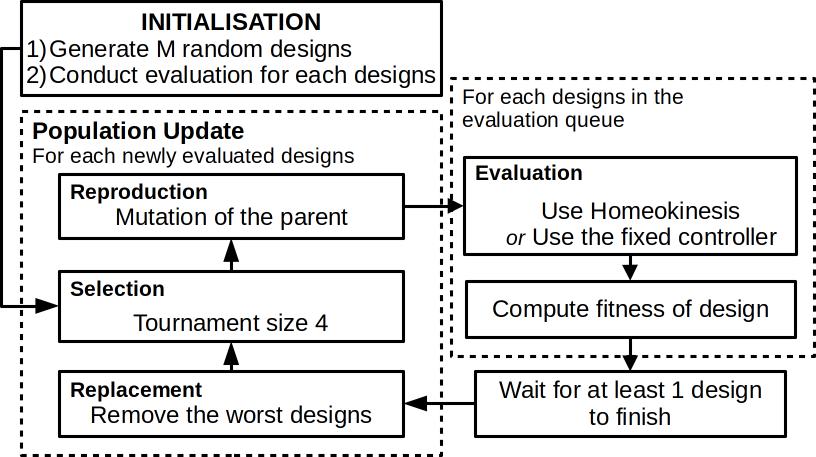}
  \caption{Asynchronous Morpho-Evolution. Our approach, MEHK, uses intrinsically motivated behaviour, homoekinesis, for evaluation.}
  \label{fig:ame}
\end{figure}

\paragraph{Morpho-evolution}
Our method is based on the Asynchronous Morpho-Evolution (AME) algorithm~\cite{le2024improving}. AME works on two sets: a population and an evaluation queue (Fig.~\ref{fig:ame}). The population is the set of robot designs already evaluated and ready for reproduction. The evaluation queue is the set of newly generated designs waiting to be evaluated. The initial population include $100$ designs produced by a randomly generated \emph{compositional pattern-producing network} for each design~\cite{stanley2007compositional} (CPPN, more details in Sec.~\ref{sec:robotic-design-space}).

At each iteration, the algorithm updates the population. Each robot design, i.e. solution or individual, is evaluated to assess their quality by calculating its fitness. In MEHK, fitness is the coverage of the robot over the environment while it is controlled by homeokinesis for $20$ minutes. As MEHK is an asynchronous EA, the population is updated as soon as a design is evaluated. This update has three steps: 

\paragraph*{\textbf{Replacement}} The newly evaluated designs are added to the population, and the same number of designs are removed, keeping the population size constant. The designs with the lowest fitness are removed.
\paragraph*{\textbf{Selection}} The selection is done with a tournament. Four designs are randomly selected from the population, and the one with the highest fitness is selected for reproduction.
\paragraph*{\textbf{Reproduction}} The topology, weights and activation functions of the CPPN that generated the selected parent are mutated.
The neurons of the CPPN can have 4 activation functions: Gaussian, sigmoid, sinusoid, or linear. The weights and activation function parameters are modified with polynomial mutation.
A forward pass of the CPPN generates a new solution.

%In MEHK, the evaluation and selection step is informed by the fast and exploratory IM homeokinetic controller. The fitness for each design is calculated in a few seconds of simulation rather than extensive learning.

% The mutation of the CPPN comprises four possible operation modifying the topology of the network: (1) add a hidden neuron and choose randomly its activation function among  function; (2) remove a hidden neuron and its connection; (3) add a connection; (4) remove a connection.  Also, the weight, biases and parameters of the activation functions are mutated using polynomial mutation. All these operations are triggered randomly biased by mutation rates specific to each operation. 

\paragraph{Homeokinetic Controller}

Homeokinesis (HK) is a principle that generates coordinated and seemingly intelligent behaviour without specific goals~\cite{der2012playful}. 
%This IM adapts the controller parameters to maintain the dynamics at the edge of chaos.
The robot's behaviour fluctuates between \emph{predictability} and \emph{sensitivity}. Predictable behaviours are the ones that result in a change in the state of the robot that can be predicted by an internal forward model. At the same time, sensitive behaviour is the one where a small action produces a significant change in the state of the robot. 
%A time-loop error is defined to update the parameters of the controller.
HK defines a controller $\mathrm C$, an internal forward model $\mathrm M$ and a sensorimotor loop $\mathbf \Psi$, with sensors $\mathbf s$ and actuators $\mathbf a$:
\begin{equation*}
\begin{split}
&\mathbf a_t = \mathrm C (\mathbf s_t),\\
&\mathbf{\tilde{s}}_{t+1} = \mathrm{M}(\mathbf{s}_t, \mathbf{a}_t),\\
&\mathbf \Psi (\mathbf s_t) = \mathrm M(\mathbf s_t, \mathrm C(\mathbf s_t)).
\end{split}
\end{equation*}
We can define the prediction error $E$ at time $t$, and the Jacobian matrix $\mathrm{L}$ as:
\begin{equation*}
\begin{split}
&E_t = \parallel \mathbf{s}_t - \mathbf{\tilde s}_t \parallel,\\
&\mathrm{L}_{ij}  = \frac{\partial \Psi_i}{\partial s_j}.
\end{split}
\end{equation*}
Thus, the time-loop error, $\mathit{TLE}$, is defined as,
\begin{equation}
    \mathit{TLE}_t = \parallel \mathrm{L}_t^{-1} E_t \parallel^2.
\label{eq:tle}
\end{equation}
The controller parameters are updated to minimise the $\mathit{TLE}$, and the forward model is updated to minimise $E$.
This definition of the $\mathit{TLE}$ assumes that an inverse of $\mathrm{L}$ exists\footnote{To calculate $\mathrm L$, an input shift $\mathbf \eta$ is applied to the sensorimotor loop as $\mathbf \Psi(\mathbf s_t + \mathbf \eta_t)= \mathbf{\tilde s}_{t+1} + E_{t+1}$, and a Taylor expansion to obtain a new form for $E_{t+1} = \mathrm{L}_t \mathbf \eta_t$}. Following Eq.~\ref{eq:tle}, we can see that the learning rule favours predictable behaviour by minimising $E$, i.e. behaviours that $\mathrm M$ predicts more accurately. At the same time, the learning rule favours sensitive behaviour by minimising the inverse of $\mathrm L$, i.e. behaviours that have a larger impact on the dynamics.
%HK bring about exploratory behaviour by self-organising two opposite forces, predictability and sensitivity.
The result is exploratory behaviour that quickly adapts to excite all the robot's degrees of freedom while maintaining control and being reactive to external perturbations.

\paragraph{Baseline: Morpho-Evolution with Fixed Controller (MEFC)}
As a baseline, MEHK is compared to AME with a fixed controller for design evaluation (Fig.~\ref{fig:ame}). The controller is a feed-forward neural network with one hidden layer of 6 neurons. The number of inputs and outputs depends on the robot design (Fig.~\ref{fig:robot}). At the beginning of an evaluation, MEFC draws weights and bias values from a uniform distribution in the interval $[-0.5,0.5]$.

% \paragraph{Morpho-Evolution with Adaptive Learning (MEAL)}
% Morpho-evolution with Adaptive Learning (MEAL) is similar to MEHK but homeokinesis is replaced by a learning algorithm optimising the parameters of an Elman network. MEAL use the MEL framework ... MEAL has two main techniques to converge faster: a controller archive for inheritance \cite{le2022morpho} and adaptive learning budget based on the learning progress \cite{le2024improving}.

\subsection{Learning Algorithm for the Downstream Tasks}

As a final step of the experiments, a selection of robots generated by MEHK are trained on a set of tasks (Sec.~\ref{sec:exp_pro}). The learning algorithm used is a variant of \textit{covariance matrix adaptation evolutionary strategies} (CMA-ES) augmented with novelty (NCMA-ES)~\cite{le2020sample}. CMA-ES is a population-based optimisation algorithm using a multivariate normal distribution to sample solutions. In our case, the solution is the parameters of an Elman network used to control the robot. 

%NCMA-ES is an iterative process where the population is evaluated on a task, then the best half of the population is used to adapt the covariance matrix and mean of the distribution. The normal distribution will progressively focus on the region of the solution space where the optimum is.

NCMA-ES has two objectives: the task score (fitness value or reward) and the novelty score. The novelty score measures how much a behaviour obtained from a solution is novel compared to the current population and past solutions. Both objectives are combined in a weighted sum. The weights are initialised to 1 for the novelty score and 0 for the task score. At each iteration, the weights are decreased and increased by the same increment ($0.05$ in our experiments). NCMA-ES progressively transitions from a pure novelty search process to a pure goal-based optimisation process.  

\subsection{Robotic design space}
\label{sec:robotic-design-space}
\begin{figure}[ht]
  \centering
  \includegraphics[width=0.75\linewidth]{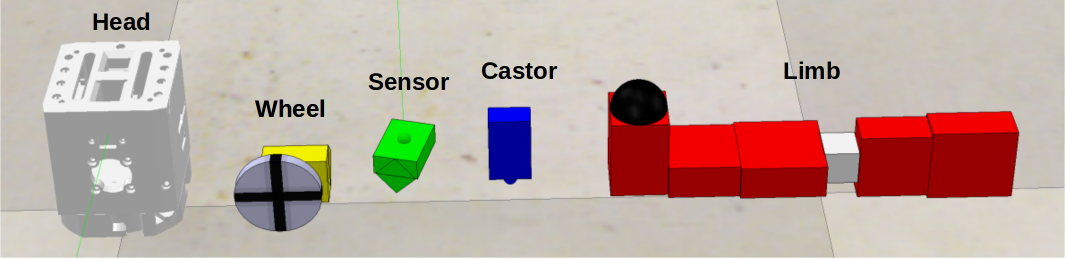} 
  \caption{Five components for the robot's design.}
  \label{fig:robot}
\end{figure}

\paragraph{Design-space}
The robotic design-space follows the ARE project~\cite{buchanan2020bootstrapping,angus2023practical}. A robot comprises 5 pre-designed components assembled on a free-formed voxel-based chassis (Fig.~\ref{fig:robot}). The chassis is limited to a matrix of voxels of $11\times11\times11$.

The \emph{head} is the central computing unit of the robots\footnote{On the physical robot, the head is composed of a battery, Raspberry Pi and custom PCBs.}. The \emph{head} is always in the middle-bottom body of the robots. A maximum of 8 components can be fixed on the surface of the chassis. There are 4 different components: limbs, wheels, sensors and castors. A limb has two degrees of freedom: one rotating on a horizontal axis and one rotating on a vertical axis. Each proximity sensor includes an infrared (IR) receiver. The proximity sensor outputs a continuous value between 0 and 1 as the distance to the closest object, and the IR receiver outputs a binary value when an object emitting IR light is detected. %With this design space, a large diversity of robots design can be generated. 

\paragraph{Controller}
The controller representation depends of which algorithm is used.

For MEHK, homeokinesis uses a pseudo-linear controller. The next action is calculated as $\mathbf a_{t+1} = g(C\mathbf s_t + \mathbf c)$, where $g$ is a sigmoidal activation function, $C$ an $m \times n$ matrix ($m$ the number of actuators, and $n$ the number of sensors), and $\mathbf c$ a bias term. 
%In all cases the number of inputs and outputs are determined by the number of actuators and sensors featured in the design, see Fig.~\ref{fig:robot}. 
 The homeokinetic forward model is defined in action-space as $\mathbf{\tilde s}_{t+1}=A\mathbf a_t + \mathbf b$, where $A$ is an $n\times m$ matrix and $\mathbf b$ is a bias term. The controller's parameters are updated with a stochastic gradient descent algorithm over the $\mathit{TLE}$ error and the model over the prediction error $E$.

For MEFC, the fixed controller is a feed-forward neural network, and NCMA-ES optimises the parameters of an Elman network to learn the downstream tasks. Both networks have 6 hidden neurons. 

For the three controllers, the input includes the angular positions of each wheel, the angular positions of each joint (2 per limb), the proximity sensor value bounded in [0,1], and IR receiver value equal to 0 or 1. The outputs are converted into target velocity for each wheel or goal position for the joints.  For MEFC and NCMA-ES, the outputs for the joints are converted into a frequency of a sinusoidal function before sending it to the robot. The input and output values for the actuators are normalised between [-1,1].
%Like the inputs, all the value outputted by the controllers are within the interval $[-1,1]$. 
%The values for the wheels are converted into a target velocity. 
%For the joints, if the controller is homeokinesis, the values are converted into a target position with a proportional controller.

\paragraph{Encoding}
The \textit{compositional pattern producing network} (CPPN)~\cite{stanley2007compositional} is used to generate the design of the robots. A CPPN neural network generates spatial patterns using spatial coordinates as inputs. The design is generated based on a 3-dimensional square matrix of size $11$. Each voxel of the matrix can be empty, be a piece of the chassis, or be the attached point of one of the four components. The CPPN is queried for each voxel to determine its content. The CPPN has four inputs: x, y, z, and r, corresponding to the 3-dimensional coordinates of the voxel and its distance to the centre of the robot. The network outputs 5 values for each voxel corresponding to each possible component: a piece of chassis, a wheel, a limb, a sensor or a castor. Details of the decoding procedure can be found in~\cite{buchanan2020bootstrapping}.
% The decoding goes as follow:
% \begin{itemize}
%     \item[] 1. The CPPN is, first, queried to form the chassis. In this phase only the first output is taken into account. 
%     \item[] 2. A "repress" mechanism is applied to remove parts of the chassis to make it realistic: remove the parts unconnected to the head and overhanging parts\footnote{these parts are difficult to 3D print.}.
%     \item[] 3. The CPPN is queried again for each voxel on the surface of the chassis to place the four possible components: sensor, wheel, limb and castor. For this phase, only the four last outputs are taken into account. 
%     \item[] 4. Finally, a second "repress" mechanism is applied to remove colliding components and components inaccessible from above\footnote{this constraint is introduce to make the robot easy to assemble.}
% \end{itemize}

% The "repress" mechanisms are used to guarantee that the design is viable in term of dynamics and also it is possible to manufacture. 

\section{Experimental Protocol}\label{sec:exp_pro}
% \begin{figure}[thpb]
%   \centering
%   \includegraphics[width=\linewidth]{figures/experiments_protocol.jpg} 
%   \caption{Experimental protocol showing the three phases and the algorithm used for each.}
%   \label{fig:exp_pro}
% \end{figure}

\begin{figure}[t!]
  \centering
  \includegraphics[width=0.4\linewidth]{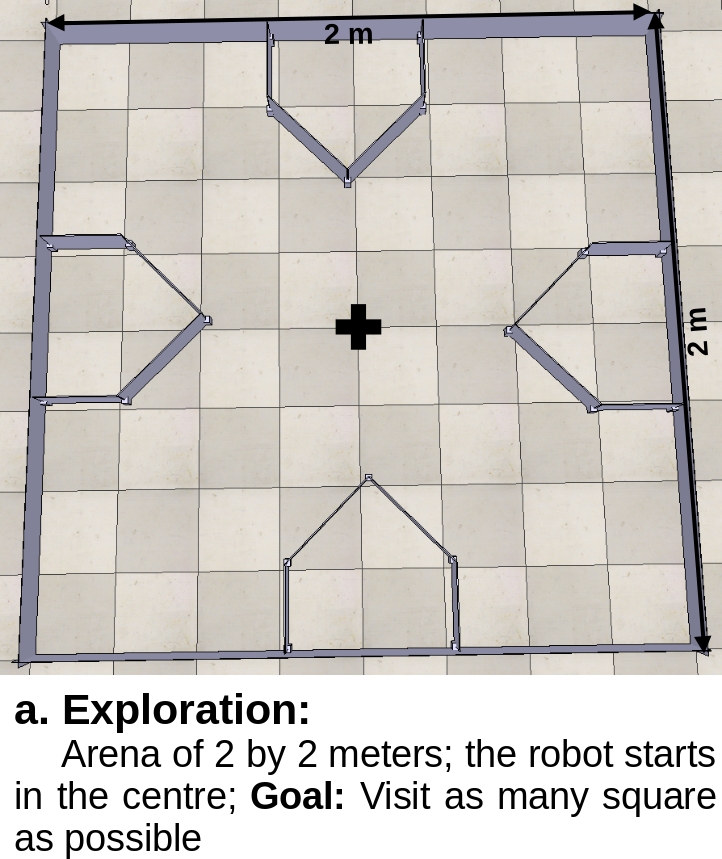}
  \includegraphics[width=0.4\linewidth]{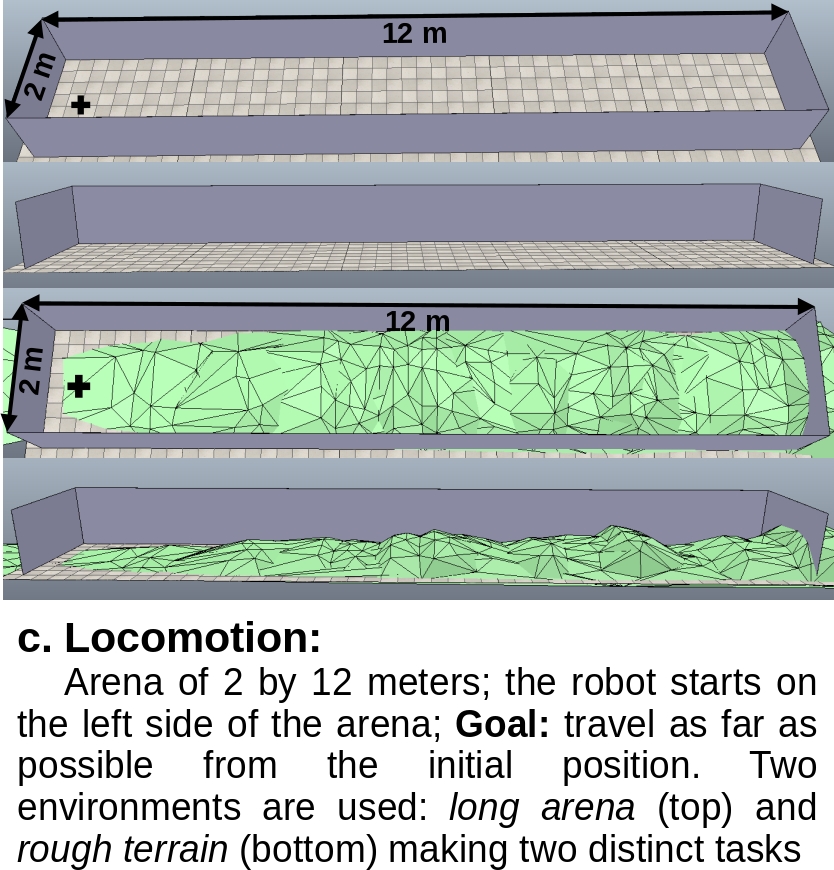}\\
  \includegraphics[width=0.4\linewidth]{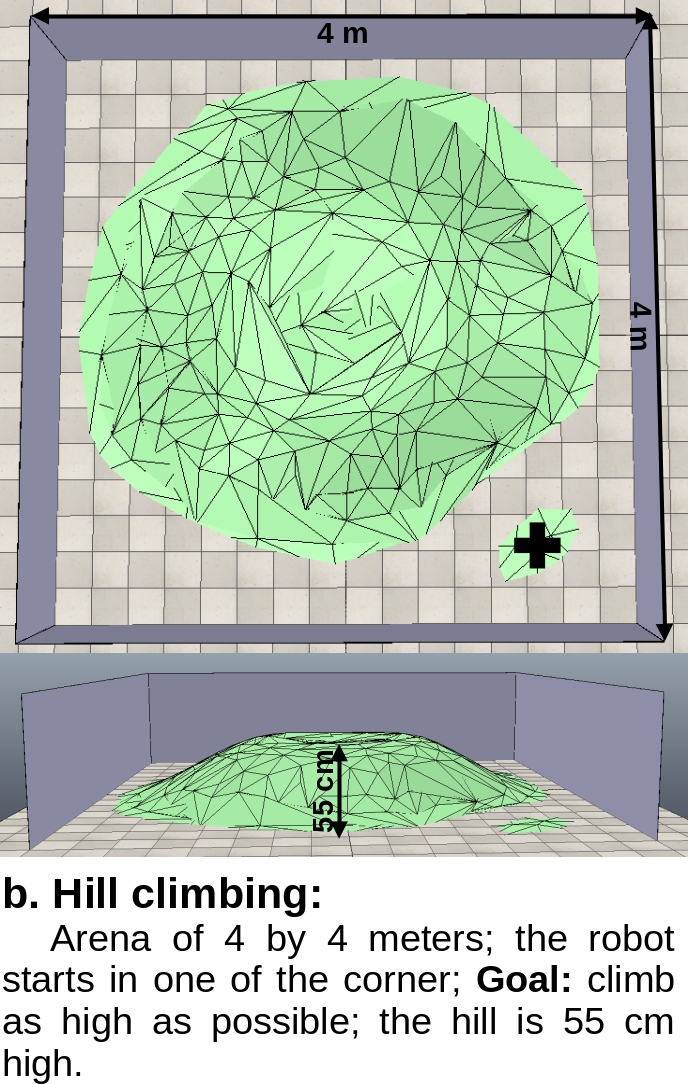}
  \includegraphics[width=0.4\linewidth]{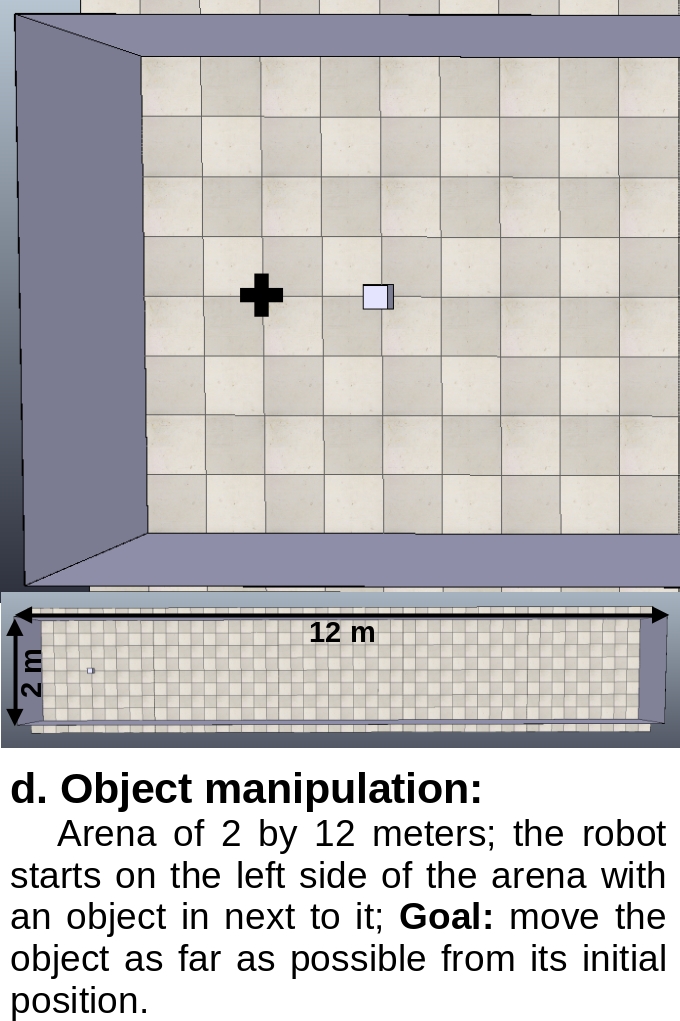}
  \caption{\textbf{a}) Arena used for the evaluation of the robots during MEHK generative phase. \textbf{b}), \textbf{c}) and \textbf{d}), are the downstream tasks environments. For clarity, \textbf{d}) only shows the beginning of the arena.}
  \label{fig:tasks}
\end{figure}

To test our approach, we first generate 10000 robot designs with MEHK and MEFC. Then, we select the best design to train them on the downstream tasks. All the experiments are replicated 30 times\footnote{The code will be available upon acceptance}.

\subsection{Generation phase}
The generation phase consists of running MEHK and MEFC on an \textit{exploration} task: the robots have 20 minutes to explore as much as possible an environment (Fig.~\ref{fig:tasks}a). The arena is divided into a grid of $8\times8$ cells to compute the exploration fitness score. The score is the number of cells uniquely visited divided by the total number of cells.
%This score is used only for the selection step of AME and not for homeokinesis which is intrinsically motivated (see section~\ref{sec:mehk}).

This phase generates diverse designs and evaluates their viability. The exploration task biases the designs for functional combinations of actuators, sensors and the shape of the chassis towards embodiment, i.e. robots able to move and interact with the environment.

%In this phase only, the proximity sensor is activated. 
The budget for each run of MEHK and MEFC is 10000 episodes, and each run of the algorithms generates 10000 designs.

\subsection{Robotic design selection}

% \begin{figure}[t!]
%   \centering
%   \includegraphics[width=0.6\linewidth]{figures/robot_selection.jpg} 
%   \caption{}
%   \label{fig:rob_select}
% \end{figure}

After the generation phase, $3$ designs are selected to be trained to solve the downstream tasks. First, a selection is made by keeping the designs with a fitness value above a set threshold (\num{0.4} for MEHK and \num{0.2} for MEFC). This selection guarantees a minimum exploratory quality of the designs.
Then, a \textit{sparsity score} based on a \textit{morphological descriptor} is computed. 

The morphological descriptor is a 3-dimensional square matrix of size $11$ representing the design-space. The matrix of integers indicates the type of components located in each voxel. $0$ indicates an empty voxel or a piece of chassis, $1$ to $4$ the presence of a component. 
To obtain a sparse matrix, the descriptor does not store the chassis, increasing the computational efficiency.
Finally, the sparsity score is computed by averaging the Euclidean distance between the design and its $15$ nearest neighbours in the descriptor space.

A Pareto front is computed based on the fitness value and sparsity score. Three designs are picked in this front: the two edges of the set, i.e. maximising the fitness value and maximising the sparsity score, and the one at the centre of the set, i.e. having a median value of fitness and sparsity.

\subsection{Downstream Tasks Learning}

The $3$ selected design are trained on $4$ tasks using NCMA-ES. 

\paragraph*{\textbf{Hill climbing}} Fig.~\ref{fig:tasks}b. Starting in one corner of a $4\times 4$ meters arena, the fitness function is the highest altitude the robot reaches after a $120$ seconds episode.
% \paragraph*{\textbf{Targeted locomotion}} Fig.~\ref{fig:tasks}c. 
% Starting in one of the corners of a $2\times 2$ meters arena, the robot has to reach an IR beacon detectable by its sensors. The robot is evaluated $3$ times with the target on different location. The fitness is the average distance between the target and the robot's final positions of each evaluation. The episodes length is $60$ seconds.
\paragraph*{\textbf{Locomotion}} Fig.~\ref{fig:tasks}c. In this task, the robot starts on the far left of a $12$ meter arena and has to travel as far as possible. The fitness function is the distance between the initial and final positions of the robot. Two environments are used for this task: an empty and a rough terrain arena. In the rough arena, the terrain gets incrementally harder to navigate. We separate the tasks into \textit{locomotion-on-flat-terrain} and \textit{locomotion-on-rough-terrain}. The episodes last 240 seconds.
\paragraph*{\textbf{Object manipulation}} Fig.~\ref{fig:tasks}d. Similar to the locomotion task, the robot starts at the left of an empty arena. A cube of $10$cm is placed $50$cm in front of the robot. The cube is an IR emitter detectable by the robot sensors. The robot has to move the cube and the fitness function is the distance between the initial and final positions of the cube. The episode's length is $240$ seconds.

We run NCMA-ES with a budget of 10000 evaluations and a population of $50$ controllers.

%\paragraph{End-to-end robotic design generation}

% \section{Implementation}

% We present the implementation of homeokinesis and the
% \subsection{Our intrinsic morphoevolution}

% Homeokinetic control equations

% Morphogenesis eqs.

% Full process.

% \subsection{Baselines comparison}

% Original Leni's morphoevolution papers.

\section{Results}

\subsection{Generation phase}

\begin{figure}[h]
  \centering
  \includegraphics[width=0.4\linewidth]{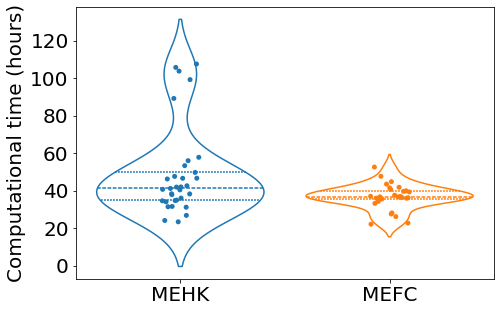}
  \includegraphics[width=0.4\linewidth]{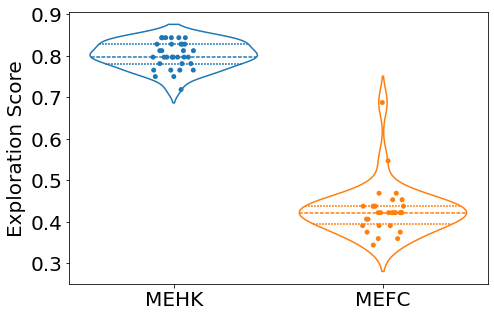}
  \caption{Comparison of MEHK (blue) and MEFC (orange). On the left, the total computational time and on the right, best exploration score obtained from each replicate. %Similar computational for MEHK and MEFC yields higher score for MEHK.
  }
  \label{fig:explo_score}
\end{figure}

Fig.~\ref{fig:explo_score} shows the distribution of the computational time and the best exploration score of MEHK and MEFC. MEHK outperform MEFC with an exploration score median of $0.8$, while MEFC has a median of $0.45$. This result shows that homeokinesis is an effective proxy for exploratory behaviour for any closed-loop robot design. In comparison, the fixed controller architecture is only effective when randomly generated parameters align with the robot's dynamics. Except for some outliers for MEHK, both variants use the same computational time with a median of around $40$ hours of wall time to generate and evaluate 10000 robots. 

\begin{figure}[t]
  \centering
  \includegraphics[width=0.49\linewidth]{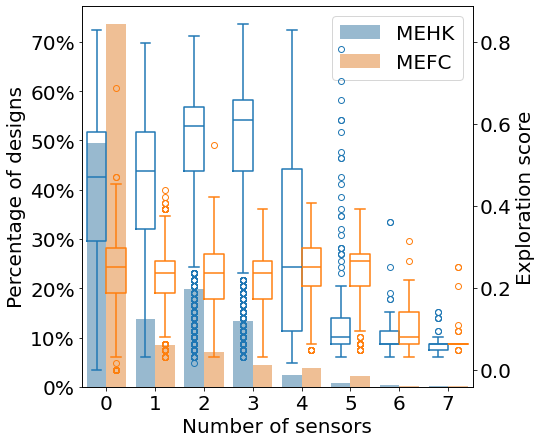}
  \includegraphics[width=0.49\linewidth]{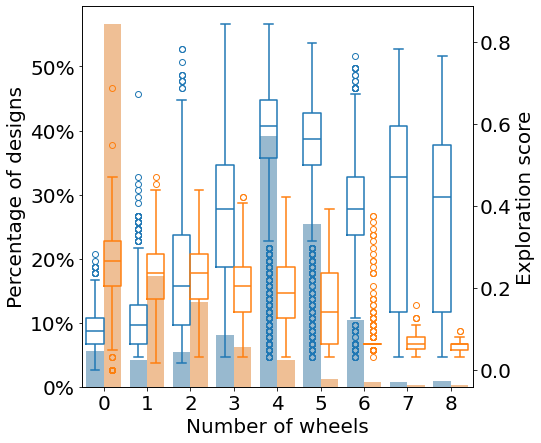}\\
  \includegraphics[width=0.49\linewidth]{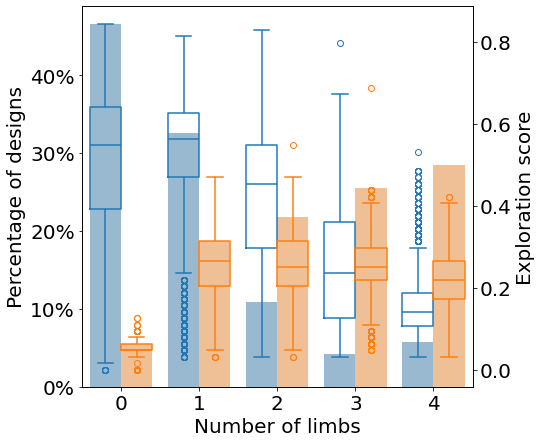} 
  \caption{Percentage of robots with respect to the number of components (filled bars), and their exploration score (empty boxes). 
  %Our approach, MEHK, shows higher score and diversity of designs compared to MEFC.
  }
  \label{fig:comp_dist}
\end{figure}

\begin{figure}[h]
  \centering
  \includegraphics[width=0.49\linewidth]{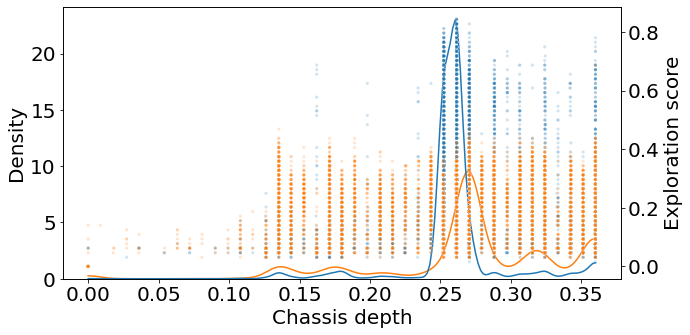}
  \includegraphics[width=0.49\linewidth]{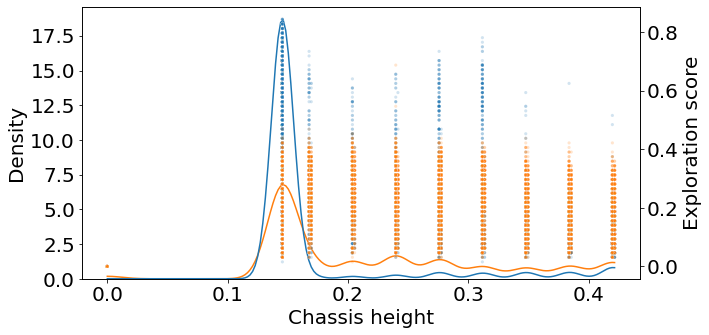}\\
  \includegraphics[width=0.49\linewidth]{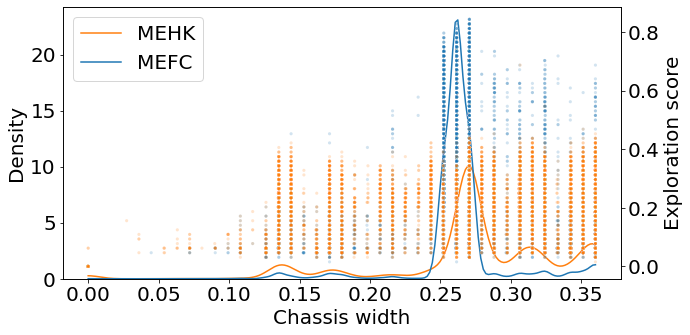}
  \includegraphics[width=0.49\linewidth]{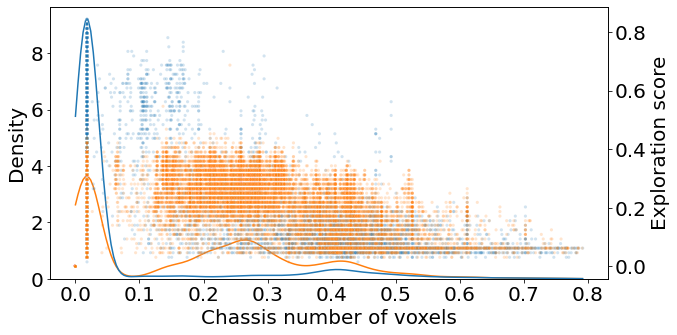}
  \caption{Exploration scores of each robot (dots) with respect to the dimension and volume of the chassis. 
  The lines represent the density distribution of the data estimated by a KDE algorithm. 
  %MEHK shows higher scores and reduced design diversity in the chassis-space compared to MEFC.
  }
  \label{fig:chas_dist}
\end{figure}

Fig.~\ref{fig:comp_dist} shows the percentage of robots generated with respect to their number of components (filled bars) and their exploration scores (empty boxes). Similarly, Fig.~\ref{fig:chas_dist} shows the exploration score over chassis depth, height and width, the number of voxels (dots) and a Kernel density estimation for the data (lines). Note that the number of sensors does not include proprioception, which is always present for wheels and limbs.
%Like for figure~\ref{fig:explo_score}, only the robots that have stayed in the population for at least one iteration of AME are included. 
%These two figures show the kind of design generated and their performances. 
The results show that half of the robots have a peak performance for $2$ and $3$ sensors, while some robots without sensors also have high performance.
MEHK generates a majority of robots with wheels ($40$\% with $4$ wheels and $30$\% with $5$ wheels) with the best exploration score.
The exploration is maximal for robots with $0$ or $1$ limb. The score decreases with the increase of limbs.
Overall, homeokinesis can efficiently control a diverse range of combinations of components but is less efficient with a larger number of limbs.
Note that the placement of the components biases the robot's behaviour, as shown by the spread of the exploration scores for different designs.

%The results are quite different for MEFC.
MEFC generates a majority of robots with at least $1$ limb ($95$\%). $55$\% of the designs have no wheel and $70$\% have no sensors. For most of the MEFC robots, the exploration score is comparable, except for lower scores for more than 5 sensors, 5 wheels, or no joints. From these results, we can see that the fixed controller can only exploit robots with joints. 

The exploration score is higher for MEHK than for MEFC. The robots generated by MEHK can explore and react to the environment more efficiently than the ones generated by MEFC. This result is explained by the nature of the behaviours that HK brings about. In this case, embodiment plays a crucial role in the designs, as the controller can only be as effective as the robot allows. At the same time, the designs evolve to enable homeokinetic behaviour. Also, Fig.~\ref{fig:comp_dist} shows a more homogeneous distribution of the number of components for the MEHK designs. MEHK can generate a wider diversity of designs than MEFC when comparing components. We believe that the improvement in exploration score and diversity is crucial for the solution of downstream tasks.

MEFC generated a higher diversity of chassis than MEHK, Fig.~\ref{fig:chas_dist}. Most MEHK designs have a small chassis that is enough to hold the head. Considering the exploration score, MEFC has a flatter distribution and is less elitist than MEHK. The bias for small chassis in MEHK makes robots more stable and less prone to falling to the side or turning upside down while exploring. Even with this bias, MEHK generated robots with high exploration scores in a range of chassis.

% \subsection{Robot Selection}

% \begin{figure}[h]
%   \centering
%   \includegraphics[width=0.49\linewidth]{figures/full_pareto.png}
%   \includegraphics[width=0.49\linewidth]{figures/pareto_selection.png}
%   \caption{\textit{Left:} Pareto front of designs generated on exploration score and morphological sparsity of all the 30 replicates combined. \textit{Right:} clipped pareto front with design having an exploration score above 0.2 for MEFC and 0.4 for MEHK. The triangles represents the robots selected for the downstream tasks learning stage.}
%   \label{fig:pareto}
% \end{figure}

% Figure~\ref{fig:pareto} shows the pareto front for the exploration score and the morphological sparsity for MEFC and MEHK. The left handside of the figure shows the full pareto front. While for MEHK, the designs are evenly distributed over the front, for MEFC the designs are clustered on the high sparsity side of the front. This confirm the low elitist nature of MEFC which produces more diversity than high performing designs. On the right handside is plotted the pareto front used for the selection where all the design bellow 0.2 for MEFC and 0.4 for MEHK are removed. The triangles represents the robots selected (3 per replicate) for the next stage of the experiments. For MEHK, the selection of robot is well distributed across the front while for MEFC the robot selected are mainly located on the high sparsity side of the pareto front. This simple selection process is able to select a range of robots by balancing the diversity and performance criteria.  

\subsection{Downstream Tasks}

\begin{figure}[h]
  \centering
  \includegraphics[width=0.9\linewidth]{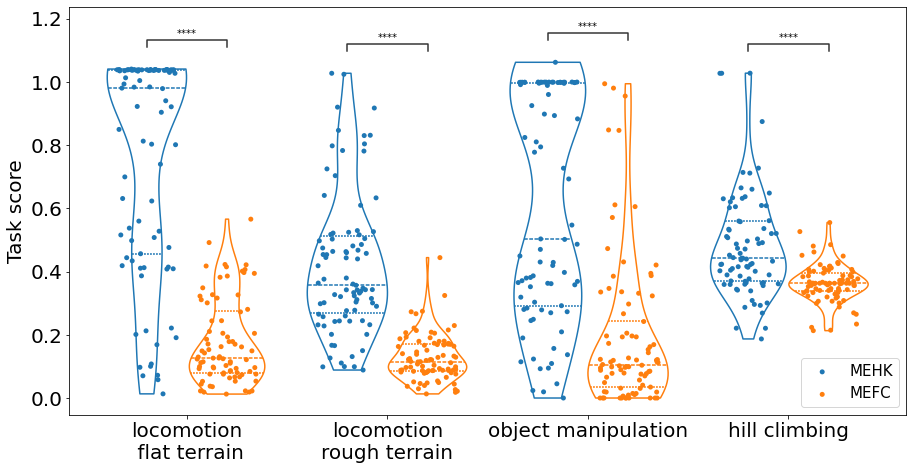} 
  \caption{Best performance scores of each robot for the different tasks. Each dot represents one robot, and the violin plots synthesise the distributions.
  %Our approach, MEHK, is able to solve all the tasks with a higher score compared to MEFC. 
  The four stars above the results indicate a p-value less than $10^{-4}$ in a Brunner-Munzel statistical test.}
  \label{fig:task_perf}
\end{figure}

\begin{figure}[h]
  \centering
  \includegraphics[width=0.49\linewidth]{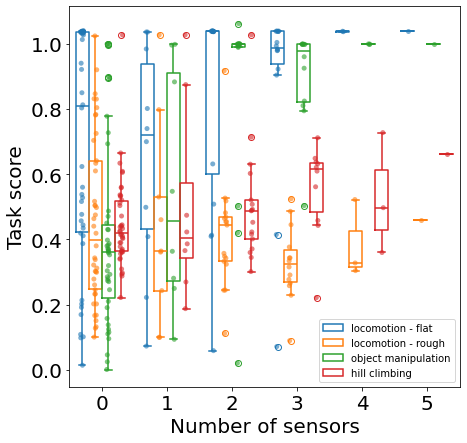}
  \includegraphics[width=0.49\linewidth]{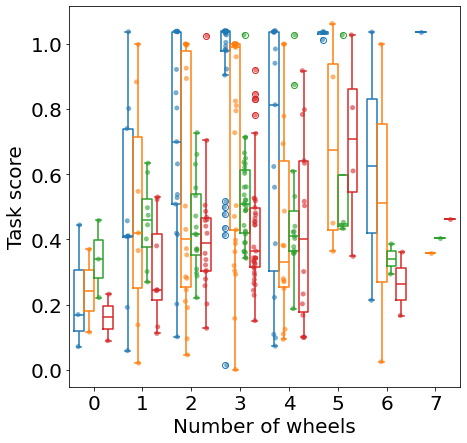}\\
  \includegraphics[width=0.49\linewidth]{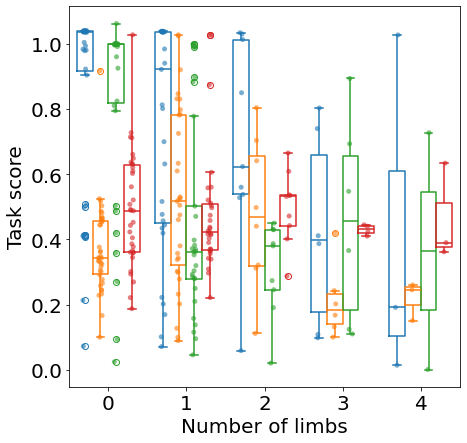}
  \caption{MEHK results for the 4 tasks over the number of sensors (top left), number of wheels (top right) and the number limbs (bottom). %Locomotion tasks are solved when wheels are present. Similarly, object manipulation is solved when sensors are present. Limbs can hinder hill climbing due to interactions with the environment and higher computational cost for NCMA-ES.
  }
  \label{fig:comp_dist2}
\end{figure}

Fig.~\ref{fig:task_perf} shows scores obtained by each robot on the downstream tasks. Maximum, median and distribution of the results show that MEHK solves all the tasks and has higher scores than MEFC.
In the flat terrain locomotion task, most MEHK robots reach the end of the arena (score greater or equal to $1$). In rough terrain, the median score is close to $0.4$, which is a result of more challenging terrain. In both terrains, MEFC shows a lower median and distribution of the score.
Fig.~\ref{fig:comp_dist2} shows the task score for the locomotion tasks (blue and orange) over the number of components for MEHK. 
As expected, on the flat terrain, the best robots have wheels. Also, the best performances are achieved with more sensors due to the ability to avoid obstacles and walls. On rough terrain, the best robots have a combination of one limb and wheels. However, with more than one limb NCMA-ES fails to find a good policy due to higher dimensionality in the search-space and the interaction between the limbs and the terrain.
%Also, the sensors does not seem to help on the rough terrain. 

In the object manipulation task, MEFC scores are similar to the locomotion task except for a longer tail with a few robots able to push the object toward the end of the arena. MEHK results are split into two groups, one with high performance ($0.6$ or greater) and another with lower performance (less than $0.5$). Fig.~\ref{fig:comp_dist2} shows that most high-performing robots have sensors. The robots can change their behaviour based on the object's position, while robots without sensors operate in an open loop. Similarly to locomotion on flat terrain, the best robots have wheels.

In the hill climbing task, the median for both MEFC and MEHK are similar (around $0.4$ and $0.5$, respectively). MEHK results have a tail of higher performance. Only three robots from MEHK were able to reach the top of the hill (robots \textbf{b} and \textbf{i} in Fig.~\ref{fig:rob_ex}). Fig.~\ref{fig:comp_dist2} shows that most combinations of components have a similar distribution, with few outliers. Only robots with $4$ or $5$ wheels show higher performance.
%Finally, targeted locomotion ...

Finally, Fig.~\ref{fig:rob_ex} shows examples of high-performing robots generated by MEHK. Robots \textbf b, \textbf e, \textbf f, \textbf h, and \textbf i are generalists able to perform well on most tasks. Robots \textbf d, \textbf g, \textbf k, and \textbf l are specialists that perform well only on $1$ or $2$ tasks. Moreover, the exploration score (purple in Fig.~\ref{fig:rob_ex}) has minimal impact on the performance of the downstream tasks. In this set of robots, the one with the best performance on all the tasks does not have the best exploration score. Exploration is necessary for a successful robot design in these tasks but is insufficient. Thus, the diversity of solutions generated by MEHK is a crucial component for improved overall performance.
%This show that a minimum exploration score of 0.4 is enough to validate the viability of a design.

%

\section{Discussion and Conclusions}
We propose a new algorithm combining morpho-evolution and homeokinesis (MEHK) to generate more efficient and diverse designs in less time than current morpho-evolution approaches. We show that MEHK solutions are viable designs for downstream tasks.
%MEHK was compared to a baseline combining morpho-evolution with a controller with static parameters, a fixed controller (MEFC), which was able to generate some diversity in term of designs but the resulting robots does not perform well on the downstream tasks. Thus, MEHK has the potential to generate diverse and viable robots. 
MEHK is computationally efficient compared to MEL frameworks. With $64$ CPUs, MEHK can generate and evaluate $10000$ designs in $40$ hours.  In comparison, Gupta et al.\cite{gupta2021embodied} use MEL with deep reinforcement learning. The authors used $1152$ CPUs to generate and evaluate $4000$ designs. 

%\textbf{MEHK a potential for high diverse and viable robots}

The generated designs are biased by the policy used to evaluate them. MEFC generates robots with several joints and diverse chassis shapes, while MEHK generates a large diversity of combinations and configurations of components with smaller chassis. In future work, we are interested in studying the influence of other intrinsic motivation architectures.
%However, as the results suggests, it is not straightforward to design a fixed controller that will evaluate properly the viability of different kind of designs.

%\textbf{Different bias on designs introduce by the type of controller or learning algorithm}

%Another interesting effect of MEHK is the predominance of wheels in the robots generated. This is most likely due to the flat terrain in the exploration environment. Homeokinesis is as much design agnostic as environment and task agnostic. So, changing the shape of the terrain will also change the kind of robots generated.

%\textbf{bias toward wheels in MEHK caused by the flat terrain in generation phase}

%\textbf{Computational time untested but comparison with literature.}

%\textbf{Future work} combination with QD. Learn a model and model based RL. Combination with DeepRL. Try other intrinsic motivation and fixed controller architecture to learn more about the biases.

\section*{Acknowledgement}
This work is funded by Edinburgh Napier Unversity and EPSRC ARE project, EP/R03561X, EP/R035733, EP/R035679.

\section*{Copyright}
This work has been submitted to the IEEE for possible publication. Copyright may be transferred without notice, after which this version may no longer be accessible.

\addtolength{\textheight}{-12cm}   % This command serves to balance the column lengths
                                  % on the last page of the document manually. It shortens
                                  % the textheight of the last page by a suitable amount.
                                  % This command does not take effect until the next page
                                  % so it should come on the page before the last. Make
                                  % sure that you do not shorten the textheight too much.

%%%%%%%%%%%%%%%%%%%%%%%%%%%%%%%%%%%%%%%%%%%%%%%%%%%%%%%%%%%%%%%%%%%%%%%%%%%%%%%%

%%%%%%%%%%%%%%%%%%%%%%%%%%%%%%%%%%%%%%%%%%%%%%%%%%%%%%%%%%%%%%%%%%%%%%%%%%%%%%%%

%%%%%%%%%%%%%%%%%%%%%%%%%%%%%%%%%%%%%%%%%%%%%%%%%%%%%%%%%%%%%%%%%%%%%%%%%%%%%%%%
% \section*{APPENDIX}

% Appendixes should appear before the acknowledgment.

% \section*{ACKNOWLEDGMENT}

% The preferred spelling of the word ÒacknowledgmentÓ in America is without an ÒeÓ after the ÒgÓ. Avoid the stilted expression, ÒOne of us (R. B. G.) thanks . . .Ó  Instead, try ÒR. B. G. thanksÓ. Put sponsor acknowledgments in the unnumbered footnote on the first page.

%%%%%%%%%%%%%%%%%%%%%%%%%%%%%%%%%%%%%%%%%%%%%%%%%%%%%%%%%%%%%%%%%%%%%%%%%%%%%%%%

\newpage

\bibliographystyle{IEEEtran}
\bibliography{IEEEabrv,bib}

\end{document}